\documentclass[10pt,twocolumn,letterpaper]{article}

\usepackage{cvpr}              

\usepackage{graphicx}
\usepackage{amsmath}
\usepackage{amssymb}
\usepackage{booktabs}
\usepackage{pifont}
\usepackage{appendix}
\usepackage{array,multirow,makecell}
\usepackage{float}

\usepackage[pagebackref,breaklinks,colorlinks]{hyperref}

\usepackage[capitalize]{cleveref}
\crefname{section}{Sec.}{Secs.}
\Crefname{section}{Section}{Sections}
\Crefname{table}{Table}{Tables}
\crefname{table}{Tab.}{Tabs.}

\def\cvprPaperID{2208} 
\def\confName{CVPR}
\def\confYear{2023}

\usepackage{microtype}
\usepackage{booktabs} 
\usepackage{multirow}
\usepackage{caption}
\usepackage{bbm}
\usepackage{setspace}
\usepackage[ruled,vlined]{algorithm2e}
\usepackage{xcolor}

\usepackage[accsupp]{axessibility}  

\def\eg{\emph{e.g}.}

\def\etal{\emph{et al}.}

\newcommand{\xmark}{\ding{55}}%

\DeclareMathOperator*{\argmin}{arg\,min}

\usepackage[pagebackref,breaklinks,colorlinks]{hyperref}

\newcommand{\smallsec}[1]{\vspace{0.2em}\noindent\textbf{#1}}

\newcommand{\methodname}[0]{MoTok\xspace}

\begin{document}

\title{Object Discovery from Motion-Guided Tokens}

\author{
{Zhipeng Bao\thanks{Work done during an internship at TRI} $^{, 1}$ \qquad Pavel Tokmakov$^{2}$ \qquad Yu-Xiong Wang$^{3}$} \\ {Adrien Gaidon$^{2}$ \qquad Martial Hebert$^1$} \\
{ $^1$CMU \qquad $^2$Toyota Research Institute \qquad $^3$UIUC}\\
  \texttt{\footnotesize \{zbao, hebert\}@cs.cmu.edu \qquad \{pavel.tokmakov, adrien.gaidon\}@tri.global \qquad yxw@illinois.edu} \\
}
\maketitle

\maketitle

\begin{abstract}
Object discovery -- separating objects from the background without manual labels -- is a fundamental open challenge in computer vision. Previous methods struggle to go beyond clustering of low-level cues, whether handcrafted (e.g., color, texture) or learned (e.g., from auto-encoders). In this work, we augment the auto-encoder representation learning framework with two key components: motion-guidance and mid-level feature tokenization. Although both have been separately investigated, we introduce a new transformer decoder showing that their benefits can compound thanks to motion-guided vector quantization. We show that our architecture effectively leverages the synergy between motion and tokenization, improving upon the state of the art on both synthetic and real datasets. Our approach enables the emergence of interpretable object-specific mid-level features, demonstrating the benefits of motion-guidance (no labeling) and quantization (interpretability, memory efficiency).
\end{abstract}


\vspace*{-4mm}
\section{Introduction}
\label{sec:intro}

Objects are central in human and computer vision. In the former, they are a fundamental primitive used to decompose the complexity of the visual world into an actionable representation. This abstraction in turn enables higher-level cognitive abilities, such as casual reasoning and planning~\cite{kahneman1992reviewing,spelke2007core}. In computer vision, object detection has achieved remarkable progress~\cite{ren2015faster,carion2020end} and is now an essential component in many applications (\eg, driving, robotics). However, these models require a large amount of manual labels from a fixed vocabulary of categories. Consequently, learning unsupervised, object-centric representations is an important step towards scaling up computer vision to the real world. 

This topic has received renewed attention recently thanks to structured generative networks with iterative inference over a fixed set of variables~\cite{burgess2019monet,greff2019multi,engelcke2019genesis,lin2020space,locatello2020object}. These methods cluster pixels in the feature space of an auto-encoder, exhibiting behavior similar to grouping based on low-level cues, such as color or texture. Hence, they are restricted to toy images with colored geometric shapes on a plain background, and fail on more complex realistic scenes~\cite{bao2022discovering}. 
\begin{figure}[t]
    \centering
    \includegraphics[width = \linewidth]{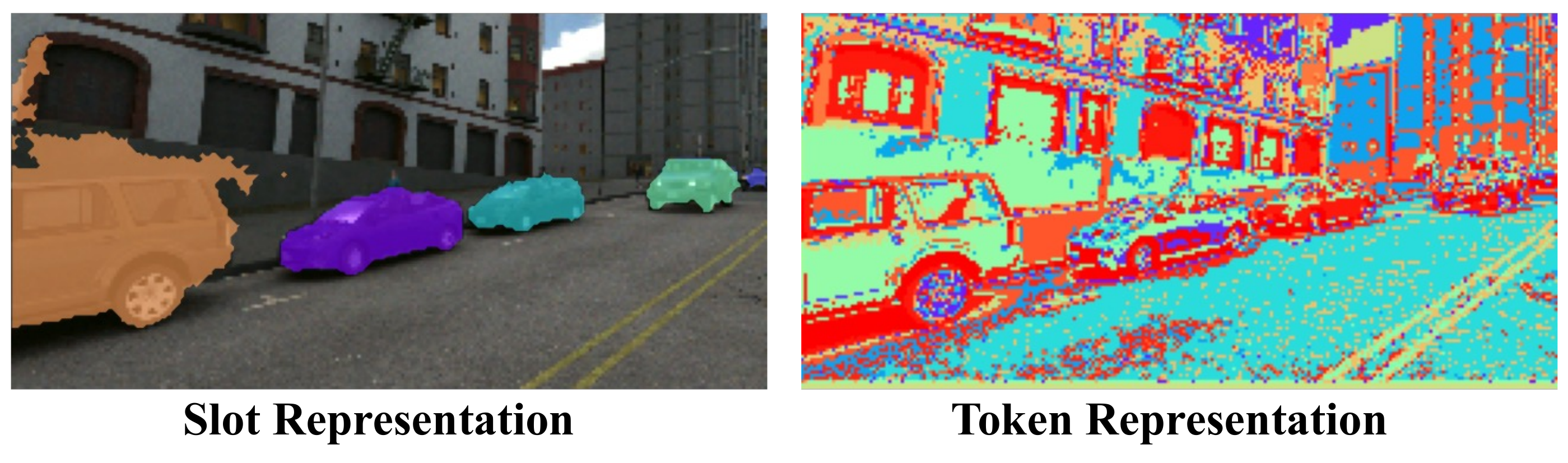}
    \vspace{-20 pt}
    \caption{TRI-PD dataset results: (left) top-10 foreground slot segments produced by our approach; (right) corresponding token representations. Compared to raw images, tokens in our framework present a more structured and compact space for reconstruction.
}
    \label{fig:teaser}
    \vspace{-10pt}
\end{figure}

Two main types of works attempt to address this shortcoming.
The first family of methods sets out to simplify the grouping problem by introducing more structure into the output space, \eg, reconstructing optical flow~\cite{kipf2021conditional} or depth~\cite{elsayed2022savi++}.
They, however, require supervision, either in the form of known poses~\cite{kipf2021conditional} or ground truth bounding boxes~\cite{elsayed2022savi++}, veering away from the unsupervised goal of object discovery.
In contrast, Bao \etal~\cite{bao2022discovering} resolve the object-background ambiguity by explicitly integrating an unsupervised motion segmentation algorithm~\cite{dave2019towards} into the pipeline, showing substantial progress on realistic scenes.
The second main direction to improve object discovery focuses on improving the decoder part of auto-encoding architectures~\cite{singh2022simple,singh2021illiterate}, replacing convolutional decoders with transformers~\cite{vaswani2017attention,ramesh2021zero} combined with discrete variational auto-encoders (DVAE)~\cite{rolfe2016discrete} to reduce memory footprint. These more sophisticated architectures improve performance without additional supervision, including on real-world sequences. However, these methods are evaluated with different protocols (metrics, datasets) and therefore no clear architectural principles have emerged yet for unsupervised object discovery. 

In this work, we introduce a novel architecture, \emph{Motion-guided Tokens (\methodname)}, based on the combination of two fundamental structural principles: motion and discretization. We define \emph{objects as discrete entities that might have an independent motion}. As prior works have shown encouraging results thanks to unsupervised motion guidance and better transformer-based decoders, we propose to \emph{leverage motion to guide tokenization}, the vector quantization process at the heart of attention mechanisms in transformer architectures (See Figure~\ref{fig:teaser}). In addition, to comprehensively evaluate the contributions of prior works, we ablate key design choices proposed in the past, such as the decoder architecture and reconstruction space, in a unified framework.

Our key contributions are as follows.
(1) We introduce a novel auto-encoder architecture, \emph{\methodname}, for unsupervised video object discovery with a new transformer decoder leveraging \emph{unsupervised motion-guided tokenization}.
(2) Our results on real and synthetic datasets show that with sufficient capacity in the decoder, motion guidance alleviates the need for labels, optical flow, or depth decoding thanks to tokenization, improving upon the state of the art.
(3) We show that our motion-guided tokens map to interpretable mid-level features, going beyond typical clusters of low-level features, thus explaining why our method scales to challenging realistic videos. 
Our code, models, and synthetic data are made available at \url{https://github.com/zpbao/MoTok/}.

\section{Related Work}
\label{sec:related}

In this work, we tackle the \textit{object discovery} problem in realistic videos with \textit{object-centric representation} by capitalizing on \textit{motion-guided tokens}. We review the most relevant works in these areas below.

\smallsec{Object Discovery} tackles the problem of separating objects from background without manual labels~\cite{bao2022discovering}. Classic computer vision methods use appearance-based perceptual grouping to parse a scene into object-like regions~\cite{koffka2013principles,felzenszwalb2004efficient,arbelaez2014multiscale}. Among them, \cite{arbelaez2014multiscale} first propose a multiscale fast normalized cuts algorithm and achieved robust object segmentation performance for static images. 

Recently, the topic of object discovery has experienced renewed attention under the name of unsupervised object-centric representation learning. A wide variety of learning-based methods has been introduced~\cite{greff2016tagger,burgess2019monet,greff2019multi,lin2020space,locatello2020object,engelcke2019genesis,veerapaneni2020entity,jiang2019scalor,yu2021unsupervised,singh2021illiterate}, usually with an encoder-decoder architecture~\cite{kingma2013auto,rezende2014stochastic}. These methods aim to learn compositional feature representations, \eg, a set of variables that can bind to objects in an image~\cite{greff2016tagger,burgess2019monet,greff2019multi,locatello2020object,engelcke2019genesis,veerapaneni2020entity,yu2021unsupervised,singh2021illiterate}, or a video~\cite{lin2020space,jiang2019scalor,elsayed2022savi++,kipf2021conditional,singh2022simple,karazija2022unsupervised,sajjadi2022object,bao2022discovering,seitzer2022bridging,wu2022slotformer}. Among them, \cite{locatello2020object} first formulate the SlotAttention framework, which is used to bind a set of variables, called slots, to image locations. The slots are then decoded individually and combined to reconstruct the image. 

However, without additional constraints, such methods tend to converge to pixel grouping based on low-level cues, such as color, and do not generalize to realistic images or videos with complex backgrounds. To address this limitation,~\cite{kipf2021conditional} and~\cite{elsayed2022savi++} extend the slot concept from static images to videos via reconstructing in the optical flow or depth space respectively. The intuition behind these methods is that this space provides stronger cues to separate the objects from the background.  Separately, Bao \etal~\cite{bao2022discovering} use motion cues to guide the slots to find moving objects and then generalize to all the objects that can move. 

In another line of work, Singh at al.~\cite{singh2022simple} show that a combination of a more powerful transformer decoder~\cite{dosovitskiy2020image} and discrete variational auto-encoder~\cite{rolfe2016discrete} can enhance the object discovery performance. Different from these works, we propose a unified architecture for object discovery that is flexible to different choices of decoders and reconstruction space. We also introduce the vector quantized features as an additional reconstruction space with motion guidance.

Finally, several recent works also leverage 3D geometry as inductive biases to enforce the learning-based models to focus on object-like regions~\cite{stelzner2021decomposing,chen2020object,du2020unsupervised,henderson2020unsupervised}. Though these methods remain limited to the toy, synthetic environments, the underlying geometric priors are orthogonal to our approach and have a great potential to be combined with our proposed method as a future direction.

\smallsec{Vector quantization} is originally developed for data compression in signal processing~\cite{gray1984vector}. More recently, \cite{van2017neural} propose the Vector-Quantized Variational Autoencoder (VQ-VAE), which learns a discrete representation of images, and models their distribution aggressively. This technique has motivated a series of investigations in solving different computer vision tasks including image synthesis~\cite{razavi2019generating,esser2021taming,gu2022vector}, video generation~\cite{sun2019videobert,yan2021videogpt}, and language modeling~\cite{ramesh2021zero,ramesh2022hierarchical,gu2022vector}, to name a few. In this work, we adopt the vector quantization technique as a mid-level representation in our object discovery framework. The intuition is that reconstructing in a structured, low-dimensional feature space rather than the high-variance color space should simplify the task of resolving the object and background ambiguity. In comparison,~\cite{singh2022simple} also introduce a mid-level feature space for object discovery in videos. However, they  use DVAE, not VQ-VAE, and are motivated primarily by improving training efficiency. Moreover, they did not explore combing vector quantization with motion cues in an end-to-end trainable framework.

\smallsec{Transformers} are originally proposed for sequence-to-sequence modeling in natural language processing~\cite{vaswani2017attention}. They use a multi-head attention mechanism, instead of the recurrent memory units, to aggregate information from the input. Recently, vision transformer (ViT)~\cite{dosovitskiy2020image} and its derivations have achieved state-of-the-art in several visual tasks~\cite{carion2020end,sajjadi2022scene,zhu2020deformable,jaegle2021perceiver,jaegle2021perceiverio,sajjadi2022object,meinhardt2022trackformer}. Among them, Perceiver IO~\cite{jaegle2021perceiverio} design a computationally efficient architecture to handle arbitrary outputs in addition to arbitrary inputs, greatly enhancing the generalizability of the transformer architecture. We employ a variant of a Perceiver module as a decoder in our object discovery framework to combine high representational power with computational efficiency.

\begin{figure*}[t]
    \centering
    \includegraphics[width = \linewidth]{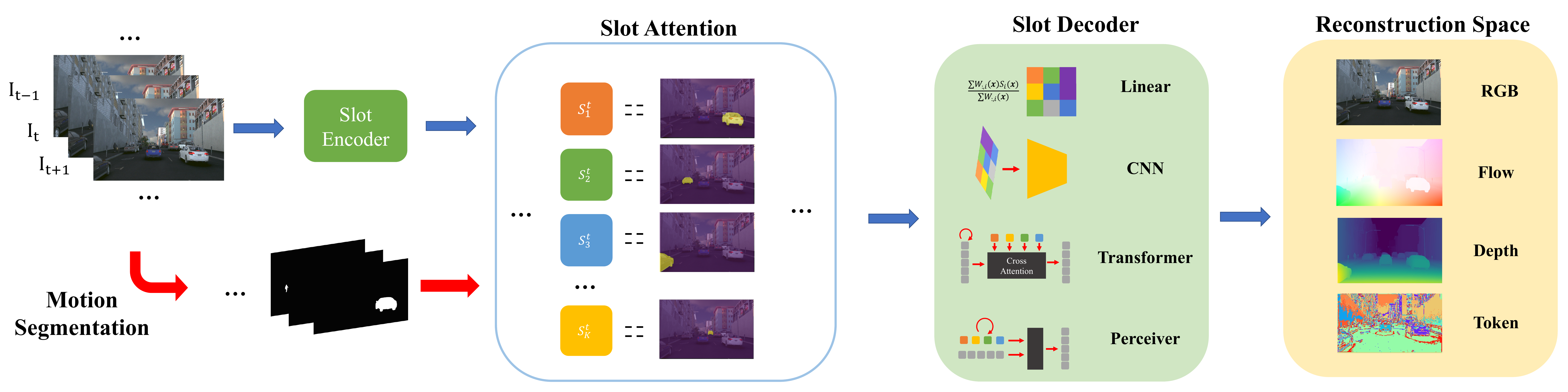}
    \vspace{-15 pt}
    \caption{Model architecture of the proposed Motion-guided Tokens (MoTok) framework. MoTok is a unified framework for video object discovery that is flexible with different choices of decoders and reconstruction spaces. Our framework effectively leverages the synergy between motion and tokenization, and enables the emergence of interpretable object-specific mid-level features.}
    \label{fig:model}
    \vspace{-8pt}
\end{figure*}
\section{Method}
\label{sec:method}

We now explain the proposed Motion-guided Tokens (MoTok) framework in detail. Its architecture is shown in Figure~\ref{fig:model}. We first introduce a motion-guided slot learning framework~\cite{bao2022discovering}  in Section~\ref{sec:preliminary}. We then describe the slot decoder units in Section~\ref{sec:slot2token}, which decode the slot features into a reconstruction space. In Section~\ref{sec:recon}, we describe the choices of reconstruction space and explain the vector-quantized reconstruction space in detail. Finally, we demonstrate how we optimize the model in Section~\ref{sec:objective}. 

\subsection{Preliminary: Motion-Guided Slot Learning}
\label{sec:preliminary}
Our object-centric representation learning module is derived  from~\cite{bao2022discovering} (Figure~\ref{fig:model}, \emph{left}). Concretely, given a sequence of video frames $\{I^{1}, I^2, ..., I^T\}$, we first process each frame through an encoder CNN to obtain an individual frame representation $H^t = f_\mathrm{enc}(I^t)$. These individual representations are aggregated by a spatio-temporal Convolutional Gated Recurrent Unit (ConvGRU)~\cite{ballas2015delving} to obtain video encoding via $H^{'t}={\tt ConvGRU}(R^{t-1}, H^t)$, where $R^{t-1} \in \mathbb{R}^{h' \times w' \times d_\mathrm{inp}}$ is the recurrent memory state. 

Next, we perform a single attention operation with $K$ slots to directly compute the slot state $S^t = W^{t} v(H^{'t})$, where $W^t \in \mathbb{R}^{K \times N}$ is the attention matrix, $N = h'\times w'$ indicates the flapped shape dimension, and $v(\cdot)$ is the value embedding function. $W^t$ is computed using the slot state in the previous frame $S^{t-1}$ and the input feature $H^{'t}$. For the first frame, we use a learnable initial state $S^0$. At the same time, for each slot $s^t_i$, we can also obtain the attention mask $W_{:,i}^t$. 

In \cite{bao2022discovering}, a motion cue is also added to guide the slots to find moving objects. A set of sparse, instance-level motion segmentation masks $\mathcal{M} = \{M^1, M^2, ..., M^T\}$ is assumed to be provided with every video, with $M^t = \{m_1, m_2, ..., m_{C^t}\}$, where $C^t$ is the number of moving objects that were successfully segmented in frame $t$, and $m_j \in \{0,1\}^{h' \times w'}$ is a binary mask. The attention mask $W^t \in \mathbb{R}^{K \times N}$, is then supervised with the motion segments. $M^t$ is also considered as a set of length $K$ padded with $\emptyset$ (no object), and a bipartite matching is found between them with the lowest cost:
\begin{equation}
    \hat{\sigma} = \argmin_{\sigma} \sum_{i=1}^K \mathcal{L}_\mathrm{seg}(m_i, W^t_{:,\sigma(i)}),
\label{eq:match}
\end{equation}
where $\mathcal{L}_\mathrm{seg}(m_i, W^t_{:,\sigma(i)})$ is the segmentation loss between the motion mask $m_i$ and the attention map of the slot with index $\sigma(i)$. Once the assignment $\hat{\sigma}$ has been computed, the final motion supervision objective is defined as follows:
\begin{equation}
    \mathcal{L}_\mathrm{motion} = \sum_{i=1}^K  \mathbbm{1}_{\{m_i \neq \emptyset\}} \mathcal{L}_\mathrm{seg}(m_i, W^t_{:,\hat{\sigma}(i)}),
\label{eq:motion}
\end{equation}
where $\mathbbm{1}_{\{m_i \neq \emptyset\}}$ denotes that the loss is only computed for the matched slots and $\mathcal{L}_\mathrm{seg}$ is the binary cross entropy.

\subsection{Slot Decoders}
\label{sec:slot2token}
The goal of the Slot Decoder is to map the slot representation $(S^t, W^t)$ to a 2D feature map $F^t$ for the reconstruction space. As shown in Figure~\ref{fig:model} (\emph{middle}), we propose four choices for the slot decoder. 

\noindent
\textbf{Linear Decoder} directly maps the slot features $S^t$ to their corresponding positions based on the attention mask $W^t$:
\begin{equation}
    F^t_\mathrm{linear}(\mathbf{x}) = \frac{\sum^K_{i=1} S^t_i(\mathbf{x}) W^t_{i,\mathbf{x}}}{\sum^K_{i=1} W^t_{i, \mathbf{x}}},
    \label{eq:linear}
\end{equation}
where $\mathbf{x}$ is an arbitrary 2D position.

\noindent
\textbf{CNN Decoder} further adds two convolutional layers to the 2D feature map formed by Equation~\ref{eq:linear}:
\begin{equation}
    F^t_\text{CNN} = \text{CNN}\left(\frac{\sum^K_{i=1} S^t_i W^t_{i,:}}{\sum^K_{i=1} W^t_{i,:}}\right).
\end{equation}
\noindent
\textbf{Transformer Decoder} decodes the feature by querying the slot representation with a 2D positional embedding through a transformer decoder:
\begin{equation}
    F^t_\mathrm{transformer} = \text{Transformer}(P, S^t, S^t),
\end{equation}
where the query $P \in \mathbb{R}^{N \times d_p}$ is a learnable positional embedding.

Compared with the previous two linear decoders, the transformer decoder further considers the global connections between the slot features and the input query, so that it will form a more powerful feature map. However, an obvious limitation is that the transformer decoder applies self-attention to the input positional query, which is 1) redundant since the positional embedding itself is learnable, and 2) not computationally efficient therefore limits the scalability of the whole model. To solve this limitation, we further propose the {\em perceiver decoder}.

\noindent
\textbf{Perceiver Decoder} is inspired by \cite{jaegle2021perceiverio}, which designs a computationally efficient architecture to handle arbitrary outputs in addition to arbitrary inputs. Different from the original architecture, we only add a self-attention layer to the slot representations, followed by a cross-attention layer for the output and positional embedding queries. The whole procedure is illustrated in Algorithm~\ref{alg1}.
\begin{algorithm}[tb]
\caption{Perceiver decoder}
   \label{alg1}
  Perceiver($S^t$, $P$): \\
   $S^t$ $\rightarrow$ Norm($S^t$) \\
   $\hat{S}^t$ = SelfAttention($S^t$) + $S^t$ \\
   $\hat{S}^t$ $\rightarrow$ Norm($\hat{S}^t$) \\
   $\Tilde{S}^t$ = MLP($\hat{S}^t$) + $\hat{S}^t$ \\
   $\Tilde{S}^t$ $\rightarrow$ Norm($\Tilde{S}^t$) \quad $P$ $\rightarrow$ Norm($P$) \\
   $F^t_\mathrm{Perceiver}$ = CrossAttention($P$, $\Tilde{S}^t$, $\Tilde{S}^t$) 
\end{algorithm}

After forming the 2D feature map $F^t$, we decode it to a reconstruction space with a CNN-based decoder, except for when vector-quantized space is used.  As we will show next, an additional decoder is redundant in this case. 

\subsection{Reconstruction Space}
\label{sec:recon}

There are four choices of reconstruction space in our framework. The plain RGB space contains the most information but also is the most complex to resolve the object/background ambiguity. The flow and depth spaces, as shown in Figure~\ref{fig:model} (\emph{right}), are more structured. Reconstructing in these spaces makes the grouping problem easier. However, there are still shortcomings, \eg~non-moving objects are not captured in the flow space and depth cannot easily distinguish between-objects that are near each other. Moreover, these two spaces are not as informative as the RGB space. In addition, we introduce the VQ-space, which is end-to-end trainable and is both structured and informative. We describe the VQ-space below.

\smallsec{Vector-quantized reconstruction space.}
Different from the other three reconstruction spaces, here we do not predict the reconstruction directly, but rather supervise the feature map $F^t$ to match the latent embedding space of VQ-VAE.

In particular, following~\cite{van2017neural}, we define a latent embedding space $E$ as the set of $M$ vectors $e_i$ of dimension $d_{vq}$, $E = \left\{ e_i \in \mathbb{R}^{d_{vq}} | i = 1, 2, \cdots, M \right\}$.
Given an input image $I^t$, VQ-VAE first processes it with an encoder to get the output $z_e^t = \text{Encoder}_\mathrm{VQ}(I^t)$, and then the discrete latent variables $z$ are calculated by a nearest neighborhood search among the discrete feature set $E$:
\begin{equation}
    z_q^t(\mathbf{x}) = e_k, \quad  \text{where}~k = \text{argmin}_j ||z_e^t(\mathbf{x}) - e_j ||_2,
\end{equation}
and $\mathbf{x}$ is an arbitrary 2D position.
The reconstructed image $\hat{I}^t$ is then decoded by $\hat{I}^t = \text{Decoder}_\mathrm{VQ}(z_q^t)$. The objective of VQ-VAE is:
\begin{equation}
\small
    \mathcal{L}_\text{VQVAE} = \log P(I^t | z_q^t) + ||sg[z_e^t] - z_q^t||_2 + || sg[z_q^t] - z_e^t ||_2,
\end{equation}
where $sg[\cdot]$ is the stop-gradient operation.

Then we use the quantized feature map $z_q^t$ as the target signal for the slot feature map $F^t$. The final objective of VQ-VAE and the VQ reconstruction is:
\begin{equation}
    \mathcal{L}_{\text{VQ}} = \mathcal{L}_{\text{VQVAE}} + ||sg[F^t] - z_q^t||_2 + || sg[z_q^t] - F^t ||_2.
    \label{eq:vqloss}
\end{equation}

\smallsec{Motion-guided token representation.} The last term in Equation~\ref{eq:vqloss}, $||sg[F^t] - z_q^t||_2$, enables the motion signal from slot learning to jointly optimize the token space through the output of the slot decoder. Furthermore, the token representation and the motion cues build a connection linked by the slot learning, thus enabling the {\em emergence of interpretable object-specific mid-level features of tokens}. In addition, reconstructing in a more compact token space also benefits the model by better utilizing the motion signal to achieve an improved slot representation and temporal consistency.

\subsection{Optimization}
\label{sec:objective}

\smallsec{Token contrastive constraint.} The goal of reconstructing in the VQ-space is that it is more compact and of lower variation compared with the RGB space. To make the VQ-space more structured, we add an \emph{optional} contrastive constraint below to the vector space, which increases the independence between latent vectors:
\begin{equation}
\mathcal{L}_\mathrm{contrastive} = ||\mathbb{I} - \text{softmax}(E \cdot E^T)||,
\end{equation}
where $\mathbb{I}$ is the identity matrix and $E \in \mathbb{R}^{N \times d_{vq}}$ is the matrix of the feature embedding space $S$. 

The final loss function is a combination of the reconstruction objective, the motion objective, and the optional contrastive constraint:
\begin{equation}
    \mathcal{L} = \lambda \mathcal{L}_\mathrm{motion} + \mathcal{L}_\mathrm{recon}+ \lambda_c \mathbbm{1}_\mathrm{VQ} \mathcal{L}_\mathrm{contrastive},
    \label{eq:contrastive}
\end{equation}
where $\lambda$ and $\lambda_c$ are weighting factors and $\mathbbm{1}_\mathrm{VQ}$ is an indicator function. For the reconstruction loss, we set $\mathcal{L}_\mathrm{recon} = \mathcal{L}_\mathrm{VQ}$ when performing reconstruction in the VQ-space. Otherwise, we use an $L_2$ loss for reconstruction in the other three spaces.

\section{Experimental Evaluation}
\label{sec:experiments}

\subsection{Experimental Settings}
\begin{figure}[t]
    \centering
    \includegraphics[width = \linewidth]{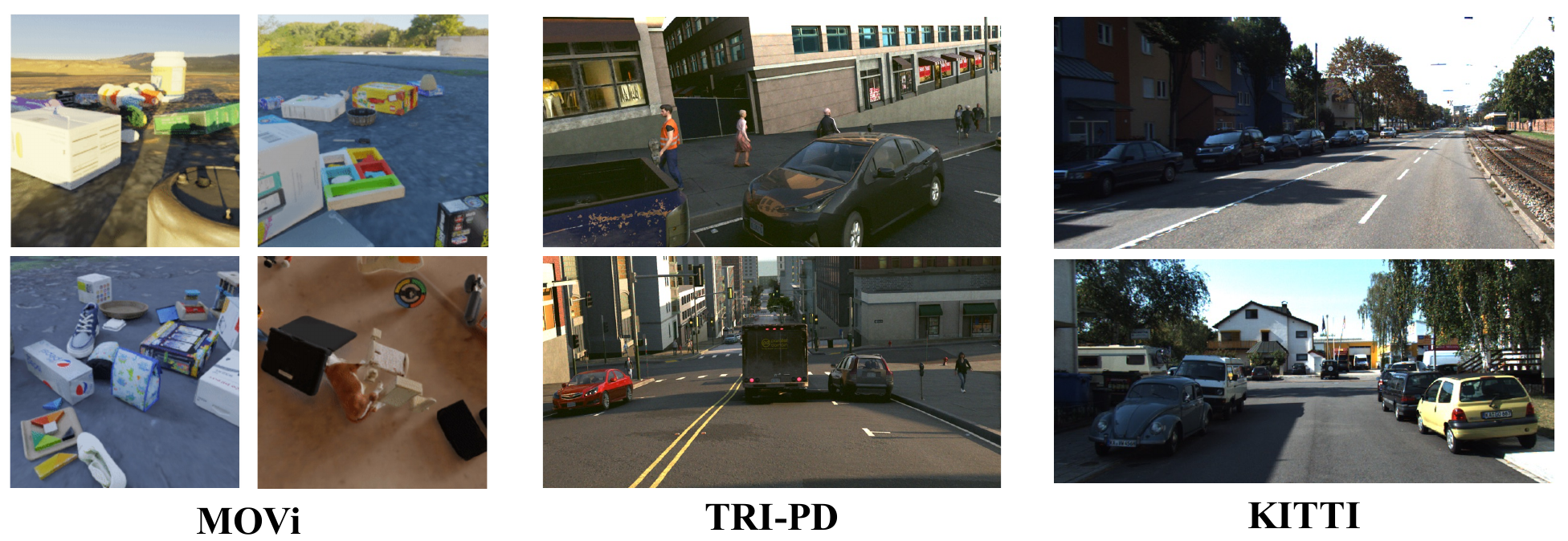}
    \vspace{-20 pt}
    \caption{Frame samples from the video datasets used in our experiments. MOVi~\cite{greff2022kubric} (left) is a multi-object video dataset created by simulating rigid body dynamics. TRI-PD~\cite{bao2022discovering} (middle) is a collection of photo-realistic, synthetic driving videos. KITTI~\cite{geiger2012we} (right) is a real-world benchmark with city driving scenes.
    }
    \label{fig:data}
    \vspace{-8pt}
\end{figure}

\smallsec{Benchmarks.} We evaluate our approach on three popular benchmarks with different complexity (Figure~\ref{fig:data}).

\noindent \textbf{MOVi}~\cite{greff2022kubric} is a synthetic multi-object video dataset, created by simulating rigid body dynamics. We use this benchmark to ablate and analyze our model architecture. Following previous works~\cite{elsayed2022savi++,singh2022simple}, we use the most complex subset, MOVi-E, for evaluation. MOVi-E contains both moving and static objects (maximum 20 objects) with linear random camera motion. The resolution of this dataset is $128 \times 128$ and each video contains 24 frames with a sample rate of 12 FPS. We use standard train and test split for MOVi-E.

\noindent \textbf{TRI-PD}~\cite{bao2022discovering,tri-packnet} is a synthetic dataset based on street driving scenarios collected by using a state-of-the-art synthetic data generation service~\cite{parallel_domain}. The dataset comes with a collection of accurate annotations (\eg~for flow or depth), which we use to ablate the impact of additional information and its quality on various models. Each video in TRI-PD is 10 seconds captured at 20 FPS. There are 924 videos in the training set and 51 videos in the test set. We crop and resize each frame to the resolution of $480 \times 968$. 

\noindent \textbf{KITTI}~\cite{geiger2012we} is a real-world benchmark with city driving scenes. We use the whole 151 KITTI videos for training and the instance segmentation subsets with 200 single-frame images for evaluation. The frames are resized to $368 \times 1248$. 

\smallsec{Baselines.} We compare our methods against the most recent learning-based object discovery models. In particular, \textbf{SAVi}~\cite{kipf2021conditional} and \textbf{SAVi++}~\cite{elsayed2022savi++} are two direct extensions of the original slot attention~\cite{locatello2020object} using optical flow and depth to facilitate object-centric learning. Alternatively, \textbf{STEVE}~\cite{singh2022simple} uses a more powerful transformer decoder to enhance the object discovery performance. \textbf{Bao \etal}~\cite{bao2022discovering}, \textbf{Karazija \etal}~\cite{karazija2022unsupervised} utilize motion cues to guide object discovery. In addition, in the appendix, we compare our approach to~\cite{seitzer2022bridging}, which is a concurrent work that relies on ImageNet~\cite{deng2009imagenet} pre-training~\cite{caron2021emerging}. For \cite{kipf2021conditional,elsayed2022savi++,singh2022simple,bao2022discovering}, we use either the official code or public implementation (see the appendix for details). Due to the lack of implementation, we reuse reported results with the two very recent approaches~\cite{karazija2022unsupervised, seitzer2022bridging}.

\smallsec{Evaluation Metrics.} Following prior works, we use the Foreground Adjusted Rand Index (FG.~ARI) to evaluate the performance of the models, which captures how well the predicted segmentation masks match ground-truth masks in a permutation-invariant fashion. Notice that FG.~ARI is measured in terms of the whole video, so that the temporal consistency of the slots is considered with this metric.

\smallsec{Implementation Details.} We use the same ResNet-18 ConvGRU encoder backbone for all the compared methods following \cite{bao2022discovering}. We set the number of slots as 24 for MOVi-E and 45 for PD and KITTI based on the maximal number of objects in these datasets. On TRI-PD and KITTI, we further downsample the images by 4 for SAVi and SAVi++ to fit the GPU memory. Notice that, the other methods also produce downsampled slot masks, leading to the evaluation under the same resolution for all the methods.
All the models are trained for 500 epochs using Adam~\cite{kingma2014adam}. During training, we train all the models with a randomly sampled video sequence of a fixed length (6, 5, 5 for MOVi-E, TRI-PD, and KITTI respectively). During the inference time, all the models are evaluated frame by frame until the end. For the FG.~ARI measurement on TRI-PD, we discard any instance labels covering an area of 0.5\% or less of the first sampled video frame following~\cite{elsayed2022savi++}. We also evaluate FG.~ARI at 5 FPS due to memory limitation. We use batch size 64 for MOVi-E, and 8 for TRI-PD and KITTI.

Our method and Bao \etal~\cite{bao2022discovering} require motion segmentation signals. We apply \cite{dave2019towards}, a powerful method pre-trained on the toy FlyingThings3D dataset~\cite{mayer2016large}, taking the ground-truth flow (MOVi-E) or RAFT~\cite{teed2020raft} flow as the input. For SAVi and SAVi++, we do not use the first-frame bounding box supervision for a fair comparison. We generate the optical flow and depth annotation for KITTI using two state-of-the-art methods, RAFT~\cite{teed2020raft} and VIDAR~\cite{tri-packnet}. We also use them for ablations on TRI-PD. More details about the hyper-parameters, training scheme, and annotation generation are provided in the appendix.

\begin{figure}[t]
    \centering
    \includegraphics[width = \linewidth]{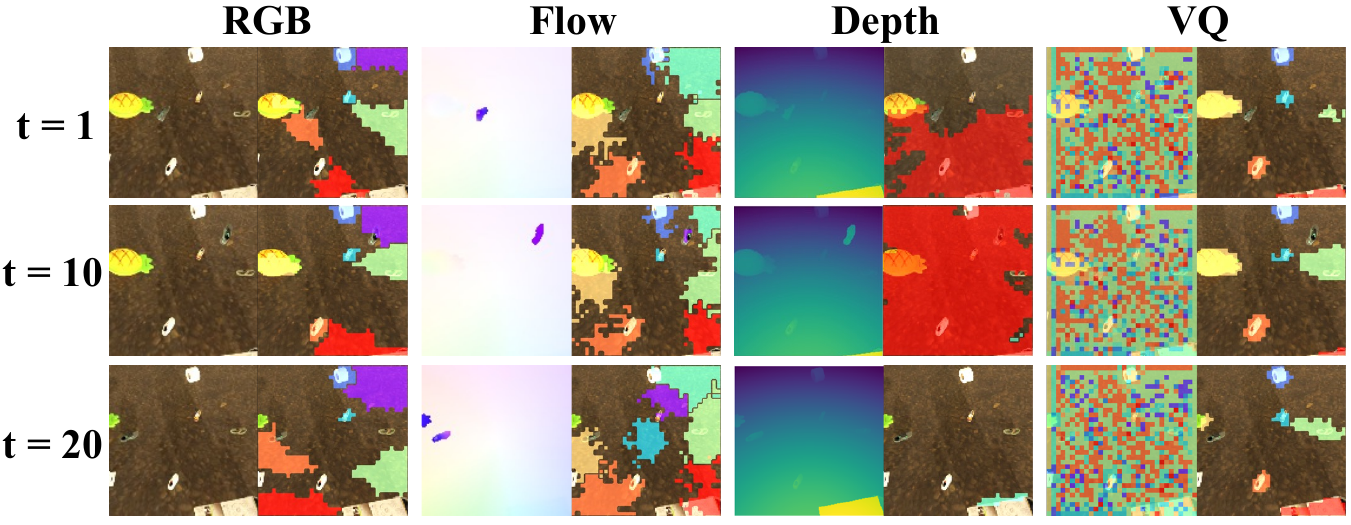}
    \vspace{-20 pt}
    \caption{Object discovery in different reconstruction spaces on the evaluation set of MOVi-E. Compared with the other reconstruction spaces, the model trained using the VQ-space demonstrates a more accurate performance and better temporal consistency.}
    \label{fig:movi}
    \vspace{-8pt}
\end{figure}

\subsection{Object Discovery on MOVi}
We use MOVi-E dataset to ablate our model with different choices of slot decoder, and reconstruction spaces, aiming to elucidate their roles in object discovery. We also compare the performance with state-of-the-art models in this section.

\begin{table}[t]
 \centering
\resizebox{0.8 \linewidth}{!}{
    \begin{tabular}{c|c|c|c}
    Motion & Space & Decoder & FG.~ARI \\ \hline
    \xmark & VQ & Linear & 14.6 \\
    \xmark & VQ & Linear-CNN & 16.3 \\
    \xmark & VQ & Transformer & 40.1 \\
    \xmark & VQ & Perceiver & {\bf 52.4} \\ \hline
    \xmark & RGB & Perceiver & 49.2 \\
    \xmark & Flow & Perceiver & 36.3 \\
    \xmark & Depth & Perceiver & 18.4 \\ 
    \xmark & Flow + Depth & Perceiver & 38.0 \\
    \xmark & VQ (flow) & Perceiver & 44.6 \\
    \end{tabular}
}
\vspace{-4pt}
    \caption{Model architecture analysis with different choices of the slot decoder and reconstruction space on MOVi-E. Perceiver decoder + VQ reconstruction space yields the best performance, indicating that (1) the capacity of the decoder plays a key role in object discovery; (2) learnable, vector-quantized reconstruction space outperforms fixed alternatives like depth or flow.}
    \label{tab:architecture}
\end{table}

\smallsec{Set up.} We first fix the reconstruction space as the VQ-space and ablate the decoder architectures (first 4 lines in Table~\ref{tab:architecture}), and then we fix the decoder architecture with the best component and reconstruct in different reconstruction space (Line 5 to 9 in Table~\ref{tab:architecture}). Since SAVi++~\cite{elsayed2022savi++} argues that reconstructing in the combined space of flow and depth is the key to their success, we add a variant \textit{Flow + Depth}, which also reconstructs in this combined space. Furthermore, as VQ-VAE can be trained to reconstruct the image in the flow space as well, we also report a \textit{VQ (flow)} variant for completeness. We do not include motion cues for the architecture analysis. A more comprehensive ablation is reported in the appendix.

\smallsec{Decoder analysis.} By comparing the results on the first four lines in Table~\ref{tab:architecture}, similar to \cite{singh2022simple}, we find that the capacity of the decoder indeed plays a key role in the object discovery performance. A simple linear decoder fails, but adding an additional CNN decoder can bring limited performance gains. Introducing powerful transformer-based decoders, on the other hand, greatly improves the object discovery capabilities of the model. Between the two transformer variants, the more advanced perceiver decoder achieves better results, while being more computationally efficient.

\smallsec{Reconstruction space analysis.} For the four reconstruction spaces, the VQ-space yields the best object discovery performance. We analyze the learned slots and token representations in Figure~\ref{fig:movi} and make the following discoveries. Firstly, the model trained using the VQ-space better separates the objects from the background and shows stronger temporal consistency. This is due to the more structured, compact, and lower-variance reconstruction space provided by the quantized features, which simplifies the task of grouping, compared to the raw RGB space. 

Secondly, by comparing \textit{Flow} and \textit{Flow+Depth} variants, we find that the latter indeed carries more information, allowing the model to better group the objects, which is consistent with~\cite{elsayed2022savi++}. However, interestingly, we also find that given a sufficiently strong decoder, reconstructing in depth or flow space does not bring further improvements compared to RGB, and can even decrease the performance. 
Finally, comparing \textit{Flow} and \textit{VQ(flow)}, we find that although adding the quantization improves performance significantly, this variant still lags behind reconstructing in the raw RGB space, reinforcing our previous conclusions.

\smallsec{Comparison to the state of the art.} In Table~\ref{tab:movi}, we compare our approach to the state-of-the-art methods. Our model achieves the best performance among them. Additionally, compared to SAVi and SAVi++, even our variants that reconstructs in the same space outperform them, indicating that the perceiver decoder and single-stage decoding strategy are the optimal choices. Our model also outperforms STEVE, due to the fact that, in contrast to DVAE, the codebook in VQ-VAE can be jointly optimized with slot learning (see Equation~\ref{eq:vqloss}). Compared with Bao \etal, our model shows better performance even without motion cues used by that method, but further incorporating them into our approach allows it to achieve top results. 

Notice that, we compare with the variants of SAVi and SAVi++ that do not rely on ground-truth bounding boxes at \textit{test time}. Their published numbers are included at the bottom of the table for reference, but they are not comparable to the other \textit{unsupervised} methods.

\begin{table}[t]
 \centering
 \resizebox{0.55 \linewidth}{!}{
    \begin{tabular}{l|c}
    Model & FG.~ARI  \\ \hline
    Bao \etal & 51.6 \\ 
    STEVE & 51.2 \\  
    SAVi & 28.1 \\ 
    SAVi++ & 31.7 \\  \hline
    MoTok (no motion) & \bf 52.4\\
    MoTok & \bf 63.8 \\ \hline
    \color{gray}SAVi (G.T. box)  & \color{gray} 53.4\\ 
    \color{gray} SAVi++ (G.T. box) & \color{gray} 84.1\\
    \end{tabular}
}
\vspace{-4pt}
    \caption{Comparison to the state-of-the-art object discovery approaches on the validation sets of MOVi-E using FG.~ARI. Our approach outperforms all the recent methods with the help of motion cues and vector quantization.}
    \label{tab:movi}
    \vspace{-10pt}
\end{table}

\begin{figure*}[t]
    \centering
    \includegraphics[width = \linewidth]{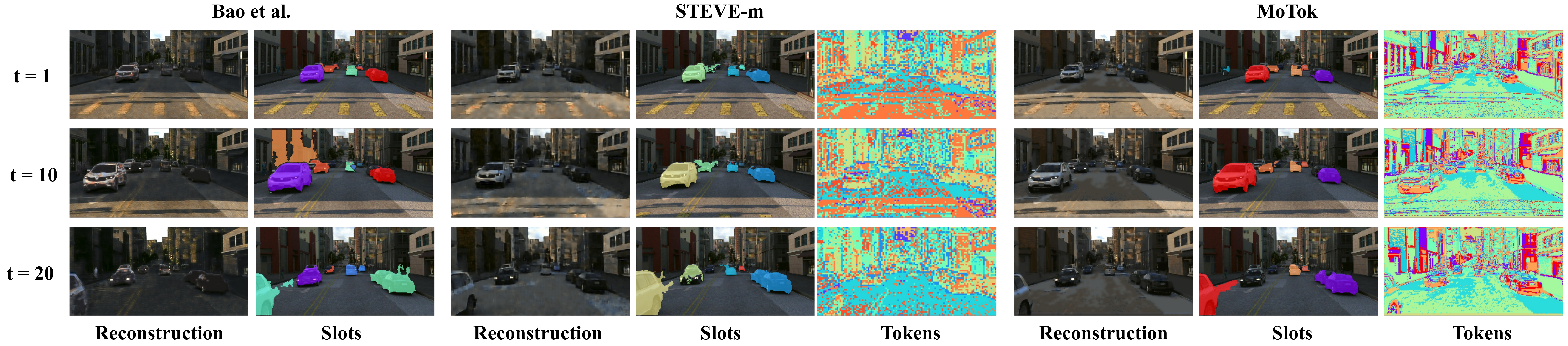}
    \vspace{-20 pt}
    \caption{Reconstruction, slot representation, and token masks for \cite{bao2022discovering}, STEVE-m~\cite{singh2022simple}, and our method. We use ground-truth motion cues for supervision, and visualize top-10 masks excluding background slots. MoTok effectively leverages the synergy between motion and tokenization, enabling the emergence of interpretable object-specific mid-level features, which simplify the problem of object discovery.}
    \label{fig:pd}
    \vspace{-12 pt}
\end{figure*}

\subsection{Object Discovery with Realistic Driving Videos}
\label{sec:pd}
\smallsec{Set up.} Firstly, following~\cite{bao2022discovering}, we capitalize on the ground-truth annotations available in TRI-PD, and evaluate the importance of the quality of external signals for various approaches. In particular, we separately evaluate using ground-truth (GT) flow, depth, and motion segmentation, as well as estimating those using state-of-the-art algorithms~\cite{dave2019towards,teed2020raft,tri-packnet}. In addition, we report a variant \textit{STEVE-m} for which we add the same motion segmentation supervision as to~\cite{bao2022discovering} and our approach for a fair comparison.

\begin{table}[t]
 \centering
\resizebox{0.8 \linewidth}{!}{
    \begin{tabular}{l|c|c|cc}
    \multirow{2}{*}{Model} & \multirow{2}{*}{Signal} & 
    \multirowcell{2}{GPU Mem.\\ (GB)} & \multicolumn{2}{c}{FG.~ARI} \\ \cline{4-5}
    & & & GT & EST \\ \hline
      Bao \etal  & Motion & 21.3 & 50.4 & 34.7\\
      SAVi~ & Flow & 17.3 & 17.9 & 14.2\\
      SAVi++ & Depth + Flow & 17.5 & 18.4 & 15.3 \\
      STEVE & - & 66.7 & 12.5 & - \\
      STEVE-m & Motion & 68.2 & 45.7 & 32.2 \\
      \hline
      MoTok & Motion & 10.9 & \bf 60.6 & \bf 55.0\\
    \end{tabular}
    }
    \vspace{-4pt}
    \caption{Object discovery evaluation of different models on TRI-PD dataset. GT denotes the ground-truth guidance given to each model and EST denotes the estimated cues. State-of-the-art models fail to work without motion cues; introducing token representation helps our model better utilize the motion signal.
    }
    \label{tab:pd}
\end{table}

From Table~\ref{tab:pd}, we can observe that our method strongly outperforms all the baselines. Additionally, we find that even with the ground-truth flow or depth, SAVI and SAVI++ do not work well for realistic videos with complex background and crowded scenes. Without motion cues, even equipped with a powerful transformer decoder, STEVE fails to work in this challenging setting as well. Finally, our model achieves better performance compared to Bao \etal~with both GT and estimated motion segments, indicating its better robustness.

\smallsec{Impact of motion-guided tokens:} In Figure~\ref{fig:pd}, we visualize the RGB reconstruction, slot representation, and token masks (if possible) for Bao \etal, STEVE-m, and our MoTok. There are several key observations. Firstly, compared with STEVE-m, MoTok enables the emergence of more interpretable mid-level features, even though both models are trained with motion cues and tokenization. This is due to the interplay between object and slot discovery in our end-to-end framework, whereas STEVE-m trains DVAE separately from the rest of the model.  

Secondly,  our model achieves the best RGB reconstruction result, even for the latter frames. The high quality of the reconstruction indicates that the model can better take advantage of the appearance signal to optimize the slot representation. Finally, our model demonstrates stronger temporal consistency compared with the baselines, thanks to the structured reconstruction space. To sum up, our architecture effectively leverages the synergy between motion cues and tokenization, enabling the emergence of interpretable mid-level features, which greatly simplifies the task of object discovery.

\begin{figure}[t]
    \centering
    \includegraphics[width = \linewidth]{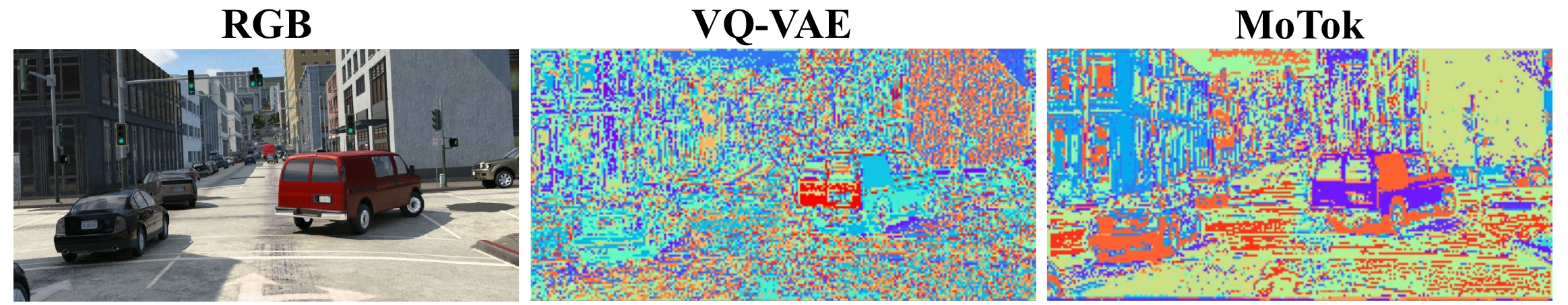}
    \vspace{-20 pt}
    \caption{Visualization of the learned tokens for vanilla VQ-VAE and MoTok. The proposed MoTok model shows a better alignment with semantic categories.}
    \label{fig:rebuttal}
    \vspace{-5pt}
\end{figure}

\smallsec{Interpretable tokens.} To better illustrate the interpretability of motion-guided tokens, we measured their alignment with the ground-truth semantic labels using cluster purity~\cite{schutze2008introduction}. Vanilla VQ-VAE (no motion guidance) achieves a purity of 40.1, whereas ours reaches 65.7 (chance performance: 38.5). Qualitative comparison in Figure~\ref{fig:rebuttal} clearly demonstrates our better semantic alignment.

\smallsec{Scalability to realistic videos:} We additionally report the GPU memory usage per batch in Table~\ref{tab:pd}. The per-slot decoding strategy and the use of transformer decoder limit the scalability of the other baselines. In comparison, our model achieves the best performance with the lowest memory consumption, showing a great generalization capability to real-world videos, thanks to the efficient perceiver decoder and the single-shot decoding strategy.

\smallsec{Impact of objectives.}  Finally, we ablate the objective design of our model with the GT annotations in Table~\ref{tab:pd_ablation}. 
We build three variants of our model, one without the additional token contrastive constraint defined by Equation~\ref{eq:contrastive} (no contrastive), the second without the motion cues (no motion), and the last trained with ground-truth instance masks for all objects to indicate the performance upper bound (*). By comparing the performance of these variants, we make the following observations: (1) in realistic synthetic videos even the strongest architectures fail to resolve the object/background ambiguity in the absence of motion cues, demonstrating the difficulty of object discovery tasks in the real world; (2) contrastive constraint facilitates diversity during vector quantization, increasing the information content of each token, which helps reduce the object/background ambiguity; (3) the performance of our model is close to the upper bound, indicating the effectiveness of the motion-guided tokens.

\subsection{Object Discovery in the Real World }
\smallsec{Set up.} To train the object discovery model on the real-world KITTI benchmark, we first pre-train all the models on PD using ground-truth flow and depth and then fine-tune them using estimated annotations on KITTI. 

\begin{figure}[t]
    \centering
    \includegraphics[width = \linewidth]{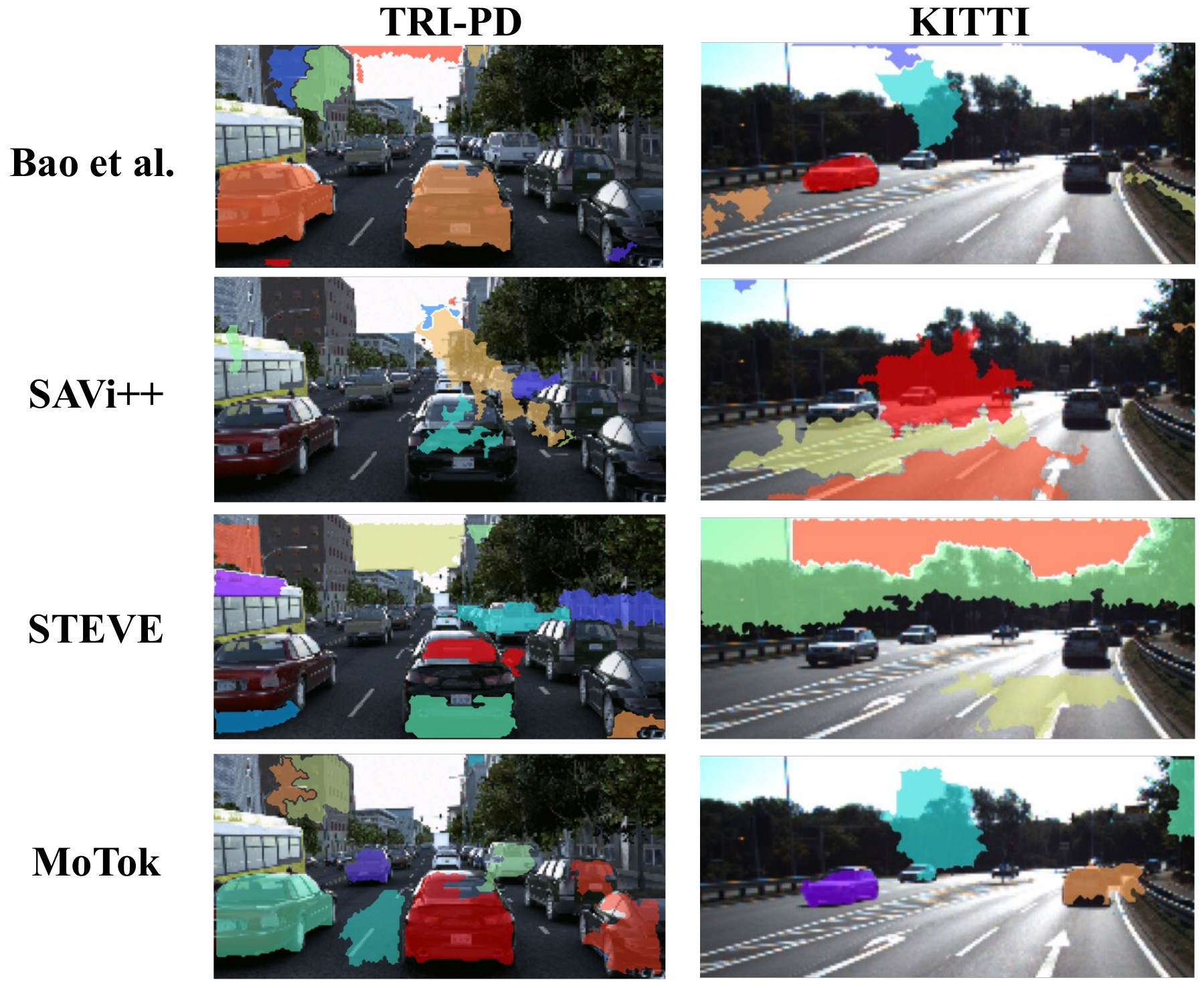}
    \vspace{-10 pt}
    \caption{Top-10 slot visualizations of all the compared methods on the evaluation set of TRI-PD and KITTI datasets. The samples on KITTI are cropped for better visualization. We discard the masks taking more than 20\% of the whole pixels. Both the more powerful decoders (STEVE) and the simpler reconstruction space (SAVi and SAVi++) fail to work in realistic driving scenes. Our model produces the most accurate masks and captures more objects. }
    \label{fig:kitti}
    \vspace{-10pt}
\end{figure}

The comparisons on KITTI are shown in Table~\ref{tab:KITTI} and Figure~\ref{fig:kitti}. We find that, consistent with our observations on TRI-PD, without motion guidance, both more powerful decoders (STEVE) and the simpler reconstruction space (SAVi and SAVi++) fail to work. However, our model still captures good dynamic object segmentation results. Compared to Bao \etal, we still achieve a better FG.~ARI score and a more solid segmentation mask, indicating the benefit of the model design and the motion-guided tokens. Finally, we outperform the very recent approach of Karazija \etal~\cite{karazija2022unsupervised} even when they include an additional warping loss objective. 

\begin{table}[t]
\begin{minipage}{0.45 \linewidth}
 \centering
\resizebox{\linewidth}{!}{
    \begin{tabular}{l|c}
    Model & FG.~ARI\\ \hline
    MoTok &\bf 60.6 \\ 
      MoTok (no contrastive) & 57.4 \\
      MoTok (no motion)
      & 15.6 \\ \hline
      $\text{MoTok}^*$  & \it 64.7 \\
    \end{tabular}
    }
    \vspace{-6pt}
    \caption{Ablation study on different learning signals. MoTok$^*$ is our model trained with ground-truth instance segmentation. The motion cue and the contrastive constraint help improve the capability of the proposed model, almost reaching the performance upper bound (MoTok$^*$).
}
\label{tab:pd_ablation}
\end{minipage}
\hfill
\begin{minipage}{0.51 \linewidth}
\vspace{-8pt}
    \centering
    \resizebox{\linewidth}{!}{
    \begin{tabular}{l|c}
    Model & FG.~ARI \\ \hline
    Bao \etal~\cite{bao2022discovering} & 47.1 \\
    SAVi & 20.0 \\
    SAVi++  & 23.9 \\
    STEVE  & 11.9 \\
    Karazija \etal~\cite{karazija2022unsupervised}& 50.8 \\
    Karazija \etal~\cite{karazija2022unsupervised} (WL)& 51.9 \\
    \hline
    MoTok & \bf 64.4\\ 
    \end{tabular}
    }
    \vspace{-6pt}
    \caption{Evaluation of object discovery in the real-world KITTI dataset. Karazija \etal~\cite{karazija2022unsupervised} (WL) is a variant trained with their proposed wrapping loss. Our model achieves state-of-the-art in real-world object discovery.
    }
    \label{tab:KITTI}
    \vspace{-5pt}
\end{minipage}
\end{table}

\subsection{Limitations and Future Work}
\smallsec{Object/background ambiguity in the real world.}
Even after successfully parsing the world into object and background slots, separating those from each other in an unsupervised way is an open challenge. Promising directions include temporal contrastive learning~\cite{jabri2020space,bian2022learning}, unsupervised clustering~\cite{caron2018deep,van2021unsupervised}, and utilizing geometric cues~\cite{zhang2019learning,zhou2022cross}.

\smallsec{Slot drift.} From Figure~\ref{fig:pd}, we notice that the final RGB reconstruction quality decreases with longer videos, which indicates that the distribution of the slot representation has shifted in the latter frames. This issue leads to degraded temporal consistency and ambiguous object masks. Better training schemes and model improvements, such as adding data augmentation during training~\cite{elsayed2022savi++}, replacing GRU units with long-short memory units~\cite{hochreiter1997long}, and enforcing temporal consistency in the token space can help.

\smallsec{Better measuring object discovery in the real world.} Measuring the performance of object discovery in the real world is challenging. As noted in~\cite{elsayed2022savi++}, one slot may re-bind to another object after the tracked object moves out of the frame and a new object enters the scene. The FG.~ARI score is not suitable for such scenarios but currently, there are no better metrics for object discovery which is able to tackle the re-binding issue. Developing a novel metric for this problem could have a significant impact on the community. 

\vspace{-2pt}
\section{Conclusion}
\label{sec:conclusion}
\vspace{-2pt}

This work proposes \textit{MoTok}, a unified architecture that leverages motion-guided tokenization for object discovery. By jointly training the slot representation with motion cues and vector quantization, our model enables the emergence of interpretable mid-level features which simplifies the problem of object discovery. Comprehensive evaluation on both synthetic and real-world benchmarks shows that with sufficient capacity of the slot decoder, motion guidance alleviates the need for labels, optical flow, or depth decoding, thanks to tokenization, achieving state-of-the-art results.

\smallsec{Acknowledgements.}
We thank Dian Chen, Alexei Efros, and Andrew Owens for their valuable comments. This research was supported by Toyota Research Institute. YXW was supported in part by NSF Grant 2106825, NIFA Award 2020-67021-32799, and the NCSA Fellows program.

{\small
\bibliographystyle{ieee_fullname}
\bibliography{egbib}
}

\clearpage
\newpage
\appendix
\setcounter{figure}{0}
\setcounter{table}{0}
\renewcommand{\thefigure}{\Alph{figure}}
\renewcommand{\thetable}{\Alph{table}}
In this appendix, we provide additional experimental results, visualizations, and implementation details that were not included in the main paper due to space limitations. We begin by showing more visualizations of our models on TRI-PD~\cite{bao2022discovering} and KITTI~\cite{geiger2012we}, including an additional sample video in Section~\ref{suppsec:visual}. The full results of the architecture analysis (Table 1 in the main paper) are provided in Section~\ref{suppsec:architecture}. In Section~\ref{suppsec:experiments}, we show additional experimental evaluations including comparisons to a most recent approach, comparisons to the state-of-the-art on MOVi-C~\cite{greff2022kubric} dataset, per-frame evaluation on MOVi-E, and comparison to STEVE equipped with a perceiver decoder. Finally, we provide further implementation details in Section~\ref{suppsec:implementation}.


\section{Additional Visualizations}
\label{suppsec:visual}

In this section, we provide more visualizations of the proposed MoToK framework on the realistic datasets -- TRI-PD~\cite{bao2022discovering} and KITTI~\cite{geiger2012we}. Full samples from the test set of TRI-PD can be found in the supplementary video, where we use the model trained with ground-truth motion cues to render the slot and token visualizations.

\subsection{TRI-PD}

\begin{figure*}
    \centering
    \includegraphics[width = \linewidth]{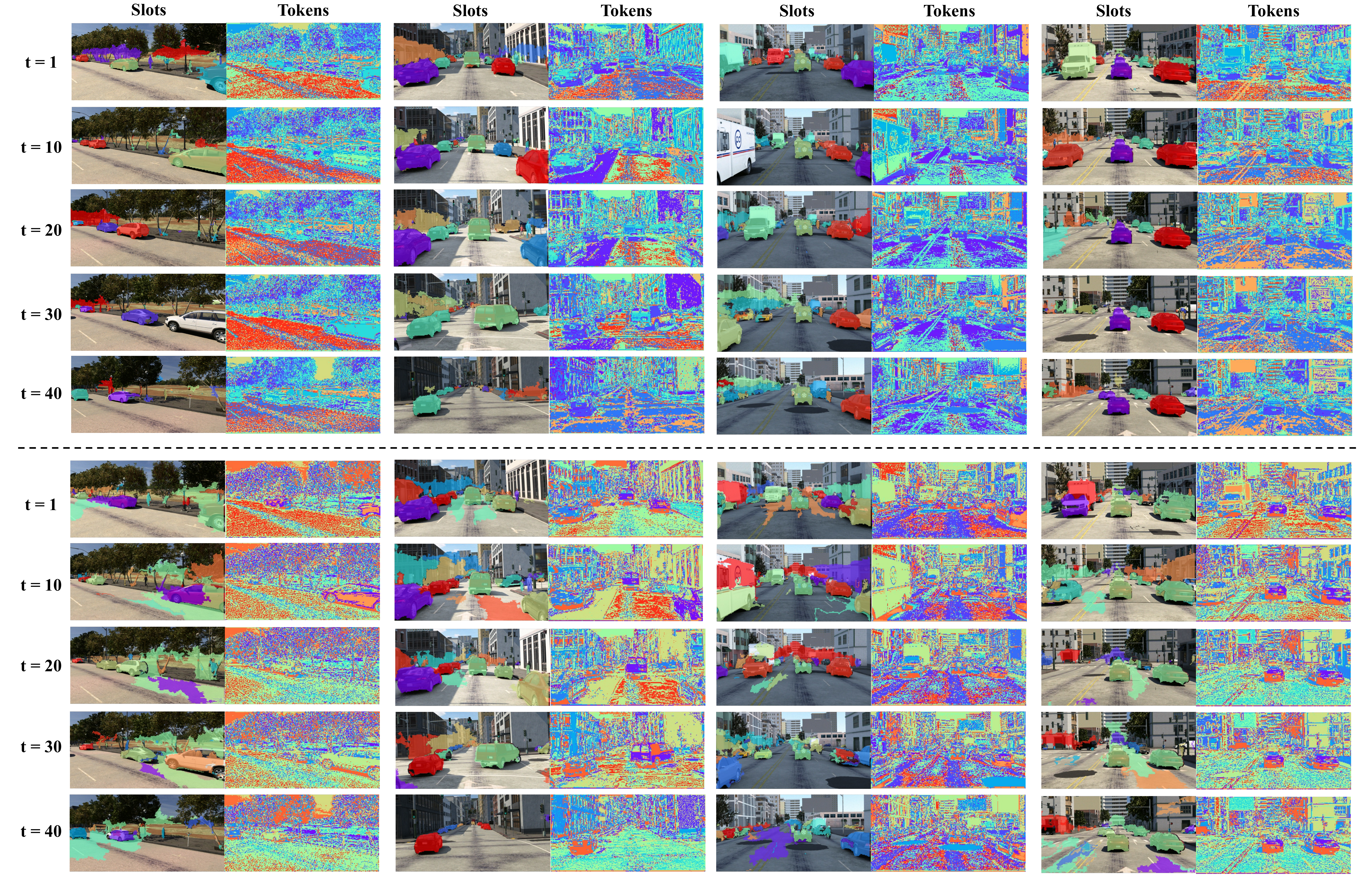}
    \caption{Slot and token visualizations of the MoTok model on TRI-PD dataset. Upper: model trained with ground-truth motion cues; bottom: model trained with estimated motion cues. Our model captures accurate object masks and achieves strong temporal consistency. Token representations simplify the grouping task in the reconstruction space by grouping pixels into mid-level regions based on textures, color, and location.}
    \label{suppfig:pd}
\end{figure*}

We show more visualizations of the discovered object masks and corresponding tokens for our model trained with both ground-truth motion segmentation (upper) and estimated using~\cite{dave2019towards} (bottom) in Figure~\ref{suppfig:pd}. Differently from the main paper, here we show the visualizations for all the slots but discard any slots that capture more than 20\% of the pixels in a frame (usually background). When trained using ground-truth motion cues, our model captures accurate object masks and demonstrates strong temporal consistency; we still maintain good temporal consistency with estimated motion but the accuracy around the object boundaries drops. 

We also find that token representations group pixels into mid-level regions based on texture, color, and location. These structured mid-level representations simplify the object discovery problem in our framework compared to grouping in the raw RGB space.

\begin{table}[t]
 \centering
\resizebox{\linewidth}{!}{
    \begin{tabular}{c|c|c|c}
    Motion & Space & Decoder & FG.~ARI \\ \hline
    \xmark & VQ & Linear & 14.6 \\
    \xmark & VQ & Linear-CNN & 16.3 \\
    \xmark & VQ & Transformer & 40.1 \\
    \xmark & VQ & Perceiver & {\bf 52.4} \\ \hline
    \xmark & RGB & Linear & 12.9 \\
    \xmark & RGB & Linear-CNN & 11.7 \\
    \xmark & RGB & Transformer & 38.4\\
    \xmark & RGB & Perceiver & 49.2 \\ \hline
    \xmark & Flow & Linear & 8.5 \\
    \xmark & Flow & CNN & 9.1 \\
    \xmark & Flow & Transformer & 35.7 \\
    \xmark & Flow & Perceiver & 36.3 \\\hline
    \xmark & Depth & Linear & 7.8 \\
    \xmark & Depth & Linear-CNN & 7.6 \\
    \xmark & Depth & Linear-Transformer & 13.2\\
    \xmark & Depth & Perceiver & 18.4 \\ \hline
    \xmark & FLow + Depth & Perceiver & 38.0 \\
    \xmark & VQ (flow) & Perceiver & 44.6 \\ \hline
    \checkmark & VQ & Perceiver & \bf 63.8 \\
    \end{tabular}
    }
    \caption{Full model architecture analysis with different choices of slot decoders and reconstruction space on MOVi-E. Perceiver decoder + VQ reconstruction space yields the best performance, indicating that (1) the capacity of the decoder plays a key role in the object discovery; (2) learnable, vector-quantized reconstruction space outperforms fixed alternatives like depth of flow.}
    \label{suptab:architecture}
\end{table}

\subsection{KITTI}

\begin{figure*}
    \centering
    \includegraphics[width = \linewidth]{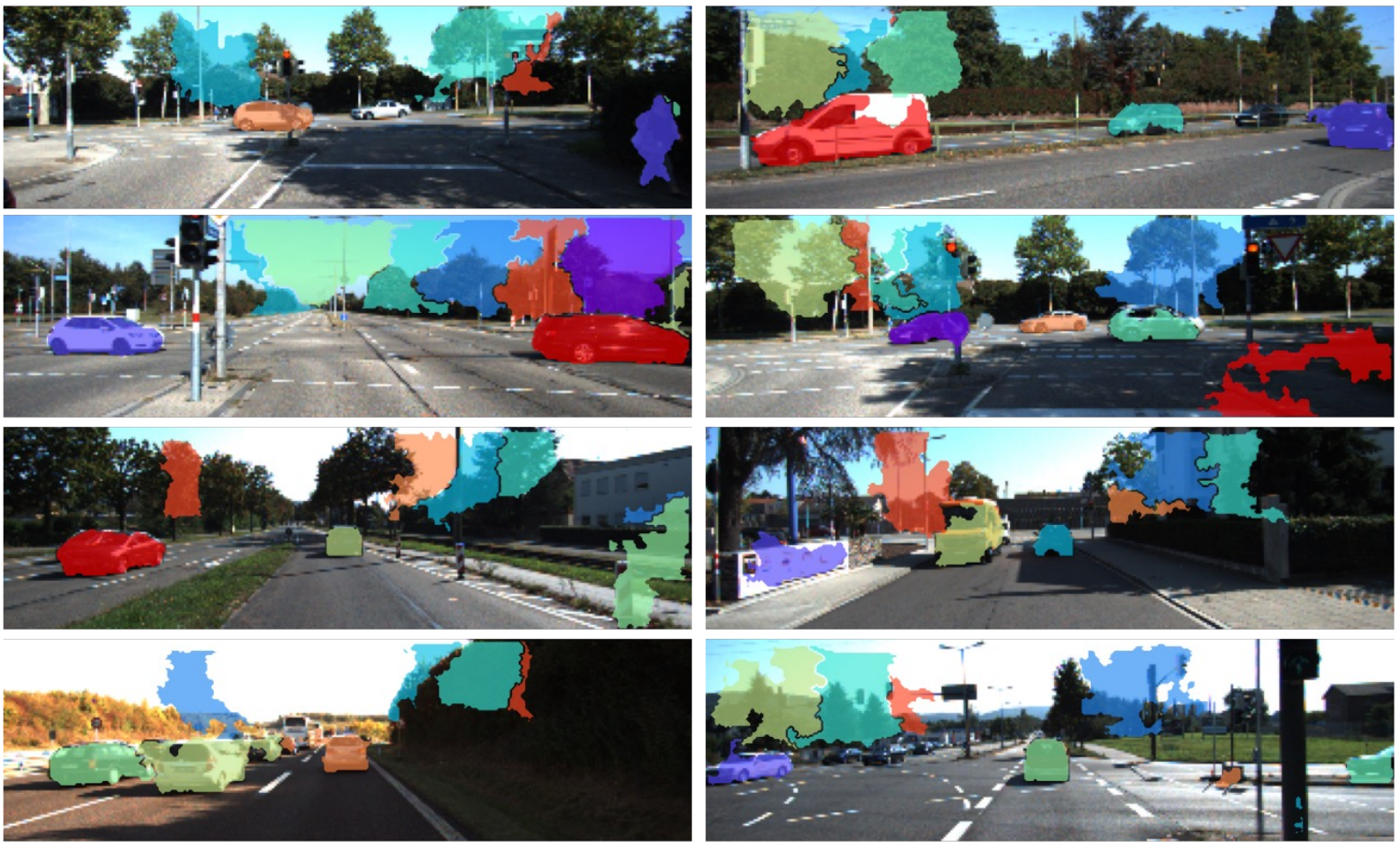}
    \vspace{-15 pt}
    \caption{Additional visualizations of our MoToK on the evaluation set of KITTI. We visualize top-10 slot masks and discard the masks containing more than 20\% of the whole pixels (usually background). Our model successfully discovers both pedestrians (top left) and vehicles (second row). However, the results are not as accurate as those of the synthetic TRI-PD, indicating the challenges of object discovery in the real world.}
    \label{suppfig:kitti}
\end{figure*}

We show additional full-resolution samples from the evaluation set of KITTI in Figure~\ref{suppfig:kitti}. Our MoTok model accurately captures both pedestrians and vehicles in the real-world in a self-supervised way. Notably, our model captures the pedestrian in the shadow in the top right, and segments all the cars in the second row, illustrating the robustness of our approach.

However, compared with the results on TRI-PD (Figure~\ref{suppfig:pd}), we miss some objects and the masks are not as precise. For example, a clearly visible car in the top left is missed, and the van in the top right is under-segmented.  These results demonstrate that object discovery in the real-world is still an open challenge, and further improvements are necessary both in the model architect and in learning objectives.

\section{Full Architecture Analysis}
\label{suppsec:architecture}

In Table~\ref{tab:architecture} in the main paper, we separately study the influence of the slot decoder architecture and the reconstruction space. In Table~\ref{suppsec:architecture}, we report the full results with the cross-selection of the slot decoders and reconstruction spaces.

Firstly, by comparing the results of different slot decoders, we find that the simple linear decoder generally fails to work. Adding an additional CNN decoder only brings limited performance gains, sometimes even hurting the performance. Between the two transformer variants, the more advanced perceiver decoder achieves top results, while being more computationally efficient, demonstrating the optimal model design of our MoToK framework. 

Secondly, the learnable VQ space yields the best object discovery performance among the four reconstruction spaces. The underlying reason is that, as shown in Figure~\ref{suppfig:pd}, the VQ space provides a structured reconstruction space, which simplifies the grouping problem. Finally, we observe that, given a powerful enough decoder, RGB space is the most effective of all the fixed reconstruction spaces we studied, and using a more structured space, such as flow and depth, can instead decrease performance.

\section{Further Experimental Evaluations}
\label{suppsec:experiments}

\subsection{Comparison to Concurrent Work}
We additionally compare our method with a concurrent approach~\cite{seitzer2022bridging}, which conducts object discovery in the ImageNet~\cite{deng2009imagenet} pre-trained feature space of DINO~\cite{caron2021emerging}. For a fair comparison, we use a ViT~\cite{dosovitskiy2020image} backbone for these experiments, but train it from scratch. In addition, following~\cite{seitzer2022bridging} we report per frame FG.~ARI. Moreover, on KITTI, we divide the whole image into 4 patches and use 9 slots for each patch. The results in Table~\ref{supptab:DINOSAUR} demonstrate that MoTok outperforms their approach on both MOVi-E and KITTI dataset even \emph{without} ImageNet pre-training, demonstrating the effectiveness of our proposed motion-guided tokenization. 

\begin{table}[t]
    \centering
    \begin{tabular}{l|cc}
        Model & MOVi-E & KITTI\\ \hline
        DINOSAUR~\cite{seitzer2022bridging} & 65.1 & 70.3 \\
        MoTok & \bf 67.8 & \bf 70.9 \\
    \end{tabular}
    \caption{Per-frame FG.~ARI comparison between DINOSAUR~\cite{seitzer2022bridging} and our MoTok on MOVi-E and KITTI datasets. We still outperform their approach on both datasets even \emph{without} ImageNet pre-training, demonstrating the effectiveness of our proposed motion-guided tokenization. }
    \label{supptab:DINOSAUR}
\end{table}

\subsection{Evaluation on MOVi-C}

We additionally evaluate our method on MOVi-C dataset~\cite{greff2022kubric}. MOVi-C features realistic foreground and background appearance. The maximum number of objects is 10, and there is no camera motion in this dataset. We compare our method with Bao \etal~\cite{bao2022discovering}, SAVI~\cite{kipf2021conditional}, SAVI++~\cite{elsayed2022savi++}, and STEVE~\cite{singh2022simple}, and \textit{MoTok (no motion)} is our model without the motion cues. 
\begin{table}[t]
 \centering
\resizebox{0.6 \linewidth}{!}{
    \begin{tabular}{l|c}
    Model & FG.~ARI \\ \hline
    Bao \etal & 59.5 \\
    SAVi &  35.1 \\
    SAVi++ & 37.4 \\
    STEVE &  49.7 \\ \hline
    MoTok (no motion) & 51.0 \\
    MoTok & \bf 69.3 \\ 
    \end{tabular}
}
\vspace{-4pt}
    \caption{Comparison to the state-of-the-art approaches to object discovery on the validation sets of MOVi-C using Fg.~ARI. Our approach outperforms all the recent methods with the help of motion cues and vector quantization.}
    \label{supptab:movic}
    \vspace{-5pt}
\end{table}

The results are shown in Table~\ref{supptab:movic}. Our model outperforms all the baselines with the help of motion cues and vector quantization. We also have the following observations. Firstly, SAVI and SAVI++, which reconstruct in the flow and depth space, achieve better performance compared with their results on MOVi-E in this scenario (all objects are moving, no camera motion).
These results indicate that, while flow and depth space can help to resolve object/background ambiguity in a simplified environment, they are not sufficient as the setting becomes more realistic. We also see that, in this dataset, motion is indeed a very strong cue as our model achieves a larger improvement from motion cues compared to MOVi-E.

\subsection{Per-Frame Evaluation on MOVi-E}

\begin{table}[t]
 \centering
\resizebox{0.7 \linewidth}{!}{
    \begin{tabular}{l|c}
    Model & per-frame FG.~ARI\\ \hline
    Bao \etal & 55.3 \\
    SAVi & 39.2 \\
    SAVi++ &  41.3 \\
    STEVE &  54.1 \\ 
    Karazija \etal~\cite{karazija2022unsupervised} & 63.1\\\hline
    MoTok (no motion) &  56.6 \\
    MoTok & \bf 66.7 \\ 
    \end{tabular}
}
\vspace{-4pt}
    \caption{Per-frame FG.~ARI evaluation on the validation sets of MOVi-E. Our approach outperforms all the recent state-of-the-art approaches with the help of motion cues and vector quantization.}
    \label{supptab:movie}
    \vspace{-10pt}
\end{table}

To compare to the reported results of~\cite{karazija2022unsupervised}, we evaluate our method on MOVi-E using per-frame FG.~ARI in Table~\ref{supptab:movie}. Our MoTok still outperforms all the recent state-of-the-art approaches with the help of motion cues and vector quantization. Comparing with the results in Table 1 in the main paper, we find that the gap between the per-frame evaluation and video-level evaluation for SAVi and SAVi++ is much larger than for the other models, showing a worse temporal consistency of these methods. These results indicate that using a more powerful decoder is also the key to achieving strong temporal consistency in object discovery.

\subsection{Equipping STEVE with the Perceiver Decoder}

Both STEVE and our model use a discrete vector space in the model design. However, we propose a more powerful perceiver decoder. To reduce the impact of the decoder, we build a baseline \textit{STEVE-p}, for which we replace the transformer decoder with the perceiver but keep the other components unchanged and report the results on MOVi-E in Table~\ref{supptab:steve}.

\begin{table}[t]
\centering
\resizebox{0.6 \linewidth}{!}{
    \begin{tabular}{l|c}
    Model & FG.~ARI\\ \hline
    STEVE  & 51.2 \\
    STEVE-p & 45.8 \\
    \hline
    MoTok (no motion) & \bf 52.4 \\ 
    \end{tabular}
    }
    \vspace{-4pt}
    \caption{Comparison to STEVE with a perceiver decoder backbone. Equipping STEVE with a perceiver decoder instead decreases the original performance. Our MoTok (no motion) still achieves better performance compared to the best variant of STEVE.}
    \label{supptab:steve}
    \vspace{-5pt}
\end{table}

Intriguingly, we find that equipping STEVE with a perceiver decoder decreases the model's performance. We analyzed their model architecture in detail and hypothesize that this is due to the fact that they use learnable DVAE token representations as the query for the transformer. Their DVAE token representations are not optimized by the DVAE reconstruction but \textit{only} optimized through the transformer decoder. Therefore, the self-attention + cross-attention mechanism in the transformer decoder is crucial to learn a rich feature representation, whereas the perceiver decoder only conducts a cross-attention operation with the query. 

In comparison, MoTok optimizes the token representation directly through VQ-VAE reconstruction, using a learned positional embedding as a lightweight query for the transformer/perceiver decoder. Therefore, the perceiver decoder leads to a better performance in the MoTok framework. Besides, our MoTok (no motion) also achieves better performance compared to the best variant of STEVE.

\section{Implementation Details}
\label{suppsec:implementation}

\subsection{MoTok Architecture Details}

We adapt the encoder and the slot attention module from~\cite{bao2022discovering}. Concretely, we first use a ResNet18 as the backbone and reduce the downsampling ratio from 16 to 4 by using stride 1 for all the convolutional blocks except for the first one (but keep the stride 2 for the first convolution \textit{layer}). We further drop the last fully-connected layers of the ResNet to obtain a feature map. We also change the hidden dimensions of the residue block to [64,64,128,128] accordingly. We use a single-layer ConvGRU with a hidden dimension of 128 and the final dimension of 64. 

For slot decoders, we use two convolutional layers for the CNN decoder with kernel sizes 5 and 3. The hidden dimensions are set to 64. For the transformer decoder and the perceiver decoder, we use both single-layer decoders, with a hidden dimension of 64 as well. We use a learnable positional embedding as the query. For the implementation of transformer and perceiver decoders, we refer to the public implementations of the transformer\footnote{\url{ https://github.com/lucidrains/vit-pytorch}} and perceiver\footnote{\url{https://github.com/esceptico/perceiver-io}}. 

For the decoder required for the flow and depth reconstruction space, we use a shallow CNN-based decoder containing 6 transposed convolutional layers. The kernel sizes are 5 except for the last layer, which is 3. The strides for the first and fourth layers are 2, otherwise are one. For the VQ space, we adopt a public VQ-VAE implementation~\footnote{\url{ https://github.com/ritheshkumar95/pytorch-vqvae}}.

We will release our code and models for reproducibility.

\subsection{Baseline Details}

\smallsec{SAVi / SAVi++~\cite{kipf2021conditional,elsayed2022savi++}} We use the same encoder backbone and slot attention architecture as in MoTok. We also add the corrector module designed by~\cite{kipf2021conditional} to the slot attention union. For the decoder architecture, we also use the same CNN decoder as ours except for the stride design. We change the strides for the first four layers to 2 and the last two layers to 1, leading to an overall up-sample ratio of 16, which is the same as the original design of SAVi and SAVi++.
We did not provide the data augmentation and the first-frame bounding box supervision to the two models. 

\smallsec{STEVE~\cite{singh2022simple}} We implement the STEVE model by referring to their paper and the public released code of \cite{singh2021illiterate}, which is a foundation work of STEVE. We use the same encoder backbone and slot attention architecture as MoTok and use the same DVAE and transformer architecture of the released code.

\smallsec{Bao \etal~\cite{bao2022discovering}} We use the public released code\footnote{
\url{https://github.com/zpbao/Discovery_Obj_Move}} for the implementation. We use the same encoder backbone and slot attention architecture as MoTok.

\smallsec{Estimated Annotation Generation} We generated estimation annotations on TRI-PD and KITTI datasets. For flow estimation, we use self-supervised SMURF flow~\cite{stone2021smurf} for the two datasets. We use a self-supervised GUDA~\cite{tri-packnet} approach to generate the depth annotations for both TRI-PD and KITTI. For the motion segmentation, we use~\cite{dave2019towards}, following the process in~\cite{bao2022discovering}.

\subsection{Training Details}
All the models are trained for 500 epochs using Adam~\cite{kingma2014adam}. Following~\cite{locatello2020object}, we use a learning rate warm-up for 3000 iterations to achieve the initial learning rate of 0.0005. For the exponential learning rate decay schedule, we set the decay rate as 0.5 and the decay step as 50,000. We set $\lambda_M$ to 1 and $\lambda_T$ to 0.05. The gradients are further clipped to a global norm value of 1, 0.05, and 0.05 for MOVi-E, TRI-PD, and KITTI respectively. We use batch size 64 for the training of MOVi-E, and 8 for TRI-PD and KITTI. We train our model with 4 NVIDIA-A100 GPUs in parallel and it will take one day, three days, and half a day to train the model on MOVi-E, TRI-PD, and KITTI accordingly. We use 128 tokens for TRI-PD and KITTI, and 64 for MOVi. See our code implementation for more details.

\end{document}